\documentclass[11pt,a4paper]{article}
\usepackage{authblk}
\usepackage[hyperref]{acl2017}
\usepackage{times}
\usepackage{latexsym}
\usepackage{xcolor,colortbl}

\usepackage[utf8]{inputenc}
\usepackage[T1]{fontenc}

\usepackage{times}
\usepackage{amsmath}
\usepackage{amssymb}
\usepackage{latexsym}
\usepackage{graphicx}
\usepackage{subcaption}

\usepackage{algorithm}
\usepackage[noend]{algpseudocode}

\makeatletter
\def\BState{\State\hskip-\ALG@thistlm}
\makeatother

\usepackage{url}

\aclfinalcopy % Uncomment this line for the final submission
\title{Data Selection Strategies for Multi-Domain Sentiment Analysis}

\author[1,2]{\bf Sebastian Ruder}
\author[2]{\bf Parsa Ghaffari}
\author[1]{\bf John G. Breslin}
\affil[1]{Insight Centre for Data Analytics}
\affil[ ]{National University of Ireland, Galway}
\affil[2]{Aylien Ltd.}
\affil[ ]{Dublin, Ireland}
\affil[ ]{\tt \{sebastian.ruder,john.breslin\}@insight-centre.org}
\affil[ ]{\tt \{sebastian,parsa\}@aylien.com}

\date{}

\begin{document}
\maketitle

\begin{abstract}

Domain adaptation is important in sentiment analysis as sentiment-indicating words vary between domains. Recently, multi-domain adaptation has become more pervasive, but existing approaches train on all available source domains including dissimilar ones. However, the selection of appropriate training data is as important as the choice of algorithm. We undertake -- to our knowledge for the first time -- an extensive study of domain similarity metrics in the context of sentiment analysis and propose novel representations, metrics, and a new scope for data selection. We evaluate the proposed methods on two large-scale multi-domain adaptation settings on tweets and reviews and demonstrate that they consistently outperform strong random and balanced baselines, while our proposed selection strategy outperforms instance-level selection and yields the best score on a large reviews corpus. All experiments are available at \texttt{url\_redacted}\footnote{Link will be made available at a later date.}

\end{abstract}

\section{Introduction}

Domain adaptation is important for sentiment analysis, as sentiment-bearing words vary between domains. If two domains are similar, such as electronics and kitchen appliance reviews, adaptation is successful; in contrast, transfer is less productive for dissimilar domains, e.g. books and electronics reviews \cite{Blitzer2007}.

Consequently, recent research has looked to the more realistic setting of multi-domain adaptation, where multiple source domains are provided and the objective is to maximize performance on the target domain. However, such approaches still train on samples from dissimilar domains that are not helpful for prediction in the actual target domain. To mitigate this, existing approaches \cite{Zhou2016} use a domain similarity measure to weight the predictions of separate source domain models. However, \citeauthor{Ruder2017a} \shortcite{Ruder2017a} show that training one model on all source data is generally more effective.

Even within one domain, such as a Twitter dataset, adaptation performance varies significantly depending on the choice of training samples \cite{Hovy2014}. In practice, data selection and domain adaptation approaches complement each other: data selection can be seen as weighting relevant instances more highly \cite{Jiang2007}, while data selection approaches in some cases have been shown to outperform adaptation methods \cite{Remus2012}.

When performing sentiment analysis in the real world, the domains are often unknown or not clearly separable. In this scenario, adaptation and data selection strategies are needed that do not presuppose a distinction of domains \cite{Plank2016}.

Data selection is also important because annotation is expensive. A large amount of unlabeled data is generally available for training, but annotation can typically only be afforded for a fraction of it. Data selection is able to guide us where to concentrate our annotation efforts.

In the following, we will review different strategies to select training data for multi-domain adaptation for sentiment analysis. For data selection, three factors are of importance: the representation, the similarity metric, and the level of the selection. With regard to the representation, we consider term distributions, word embeddings, and autoencoder representations\footnote{Note that the latter two have not been used for data selection as far as we are aware.}. We consider three domain similarity metrics: Jensen-Shanon divergence, cosine similarity, and proxy $\mathcal{A}$ distance. We finally employ three different data selection levels: domain level, training instance level, and instance subset level.

Our contributions are the following:

\begin{enumerate}
 \item We present -- to our knowledge -- the first extensive study of data selection strategies for the task of sentiment analysis.
 \item We consider novel representations and metrics for data selection and present strategies that consistently outperform random and balanced baselines on two large-scale multi-domain sentiment analysis datasets.
 \item We propose a data selection method that is well-suited to the realistic setting of unknown or ill-defined domains and consistently outperforms instance-level selection.
 \item We present guidelines for data selection for sentiment analysis in the wild.
\end{enumerate}

%Multi-domain Adversarial Autoencoder 

\section{Related work} 

\textbf{Domain adaptation.} Domain adaptation has a long history of research: \citeauthor{Blitzer2006} \shortcite{Blitzer2006} proposed a structural correspondence learning algorithm. \citeauthor{DaumeIII2007a} \shortcite{DaumeIII2007a} introduced a kernel function that maps source and target domain data to a space that encourages in-domain similarity, while \citeauthor{Pan2010a} \shortcite{Pan2010a} proposed a spectral feature alignment algorithm to align domain-specific words into meaningful clusters. \citeauthor{Glorot2011a} \shortcite{Glorot2011a} employed stacked Denoising Autoencoders to extract meaningful representations, which \citeauthor{Chen2012} \shortcite{Chen2012} extended to a marginalized version to address their high computational cost. \citeauthor{Zhuang2015} \shortcite{Zhuang2015} proposed to use deep auto-encoders for transfer learning, while \citeauthor{Zhou2016} \shortcite{Zhou2016} transferred the source examples to the target domain and vice versa using Bi-Transferring Deep Neural Networks.

\textbf{Multi-domain adaptation.} $\:$ For domain adaptation from multiple sources, \citeauthor{Mansour2009a} \shortcite{Mansour2009a} proposed a distribution weighted hypothesis with theoretical guarantees. \citeauthor{Duan2009} \shortcite{Duan2009} proposed a method to learn a least-squares SVM classifer by leveraging source classifiers, while \shortcite{Chattopadhyay2012} \citeauthor{Chattopadhyay2012} assign pseudo-labels to the target data. \citeauthor{Yang2015d} 
\shortcite{Yang2015d} embed meta-data attributes of domains, which, however, are not available when domains are unknown. \citeauthor{Wu2016a} \shortcite{Wu2016a} exploit general sentiment knowledge and word-level sentiment polarity relations for multi-source domain adaptation, while \citeauthor{Ruder2017a} \shortcite{Ruder2017a} use a distillation approach to adapt knowledge from source domain teachers to a target domain student. 

\textbf{Domain similarity metrics.} \citeauthor{Blitzer2007} \shortcite{Blitzer2007} show that proxy $\mathcal{A}$ distance can be used to measure the adaptability between two domains in order to determine, which examples should be annotated. \citeauthor{VanAsch2010} \shortcite{VanAsch2010} find that Rényi divergence outperforms other metrics in predicting POS tagging accuracy on a target domain. \citeauthor{Plank2011} \shortcite{Plank2011}, however, find that instance-level based selection on Jensen-Shannon divergence and topic models performs best for parsing data selection.

In contrast, we consider different levels of data selection and show that both selecting the most similar domain and the most similar subsets outperform instance-level selection for sentiment analysis. \citeauthor{Remus2012} \shortcite{Remus2012} also use Jensen-Shannon divergence to select training examples for sentiment analysis. Finally, \citeauthor{Wu2016a} \shortcite{Wu2016a}  propose a domain similarity metric based on a sentiment graph. Their measure, however, is only applicable if domains are clearly defined or known in advance, which is often not the case in the real world.

\section{Data selection strategies}

\subsection{Representations}

\textbf{Term distributions.} $\:$ The relative frequency distributions of terms in the vocabulary have been successfully used to gauge similarity with respect to a target domain \cite{Plank2011,Wu2016a}. The underlying assumption is that similar domains have more terms in common than dissimilar domains. The term distribution of a domain $\mathcal{D}$ is a vector $t \in \mathbb{R}^{|V|}$ where $t_i$ is the probability of the $i$-th word in the vocabulary $V$ appearing in $\mathcal{D}$. Term distributions, however, only capture superficial occurrence statistics, which likely cannot express a more nuanced spectrum of domain similarity.

\textbf{Word embeddings.} $\:$ Word embeddings have been used to capture more fine-grained notions of similarity both of words \cite{Mikolov2013d} and of sentences \cite{Wieting2016}, but have not been considered for modeling domain similarity. In line with previous work, we use a weighted sum of pre-trained word embeddings to represent each document. In particular, we discount frequent words by weighting the word embedding $v_{w_i}$ of each word $w_i$ occurring in the document $d$ with the word's smoothed inverse probability $\sqrt{\frac{a}{p(w_i)}}$ where $p(w_i)$ is the probability of $w_i$ appearing in domain $\mathcal{D}$ and $a$ is a smoothing factor, which we set to $10^{-5}$ \cite{Mikolov2013d}. The representation of a domain is then simply the mean of its document representations.

\textbf{Autoencoder representations.} $\:$ Denoising autoencoders have been successfully used in recent work on domain adaptation \cite{Glorot2011a,Zhuang2015}. Their representations are typically created to be domain-invariant, but might still capture information that is beneficial for modeling domain similarity. We train a denoising autoencoder on the data of all domains and extract the representation for each document. To obtain the representation of a domain, we take the mean of its document representations.

\textbf{Other representations.} $\:$ In the past, topic distributions have also been used and been found to be successful for part-of-speech tagging \cite{Plank2011}. While topic distributions have proven to be convincing features for sentiment analysis \cite{Lu2011}, in our experiments, we have found them not to be effective for selecting suitable training examples. We attribute this to the fact that topical similarity only inadequately captures the nuances that make up the notion of similarity for sentiment analysis.

\subsection{Domain similarity metrics}

\begin{figure*}[!htb]
    % \caption{Global caption}
    \begin{subfigure}{.32\linewidth}
      \centering
         \includegraphics[width=0.9\linewidth]{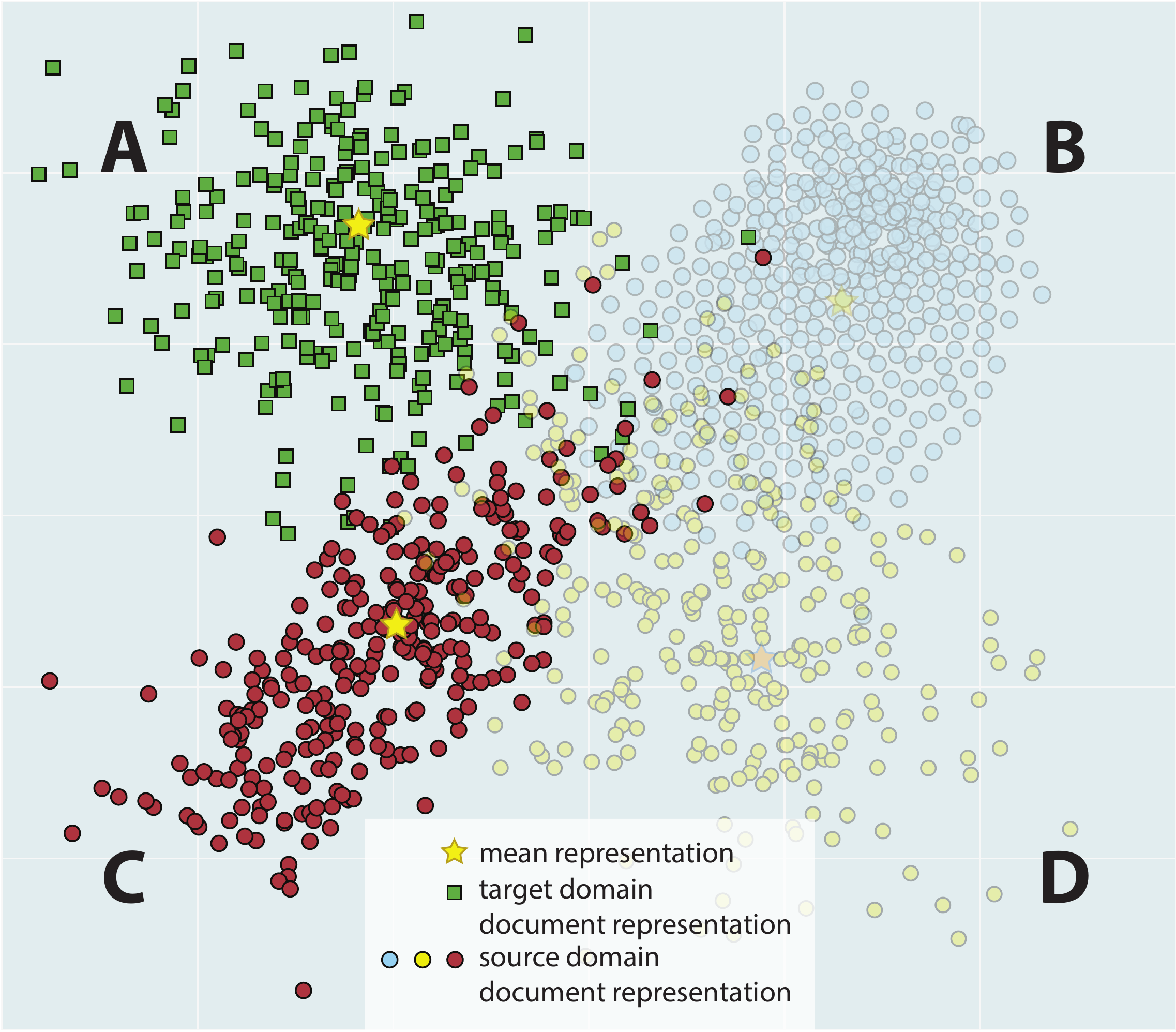}
    \caption{Domain-level selection} \label{fig:domain_level}
    \end{subfigure}%
    \hspace*{0.1cm}
    \begin{subfigure}{.32\linewidth}
      \centering
         \includegraphics[width=0.9\linewidth]{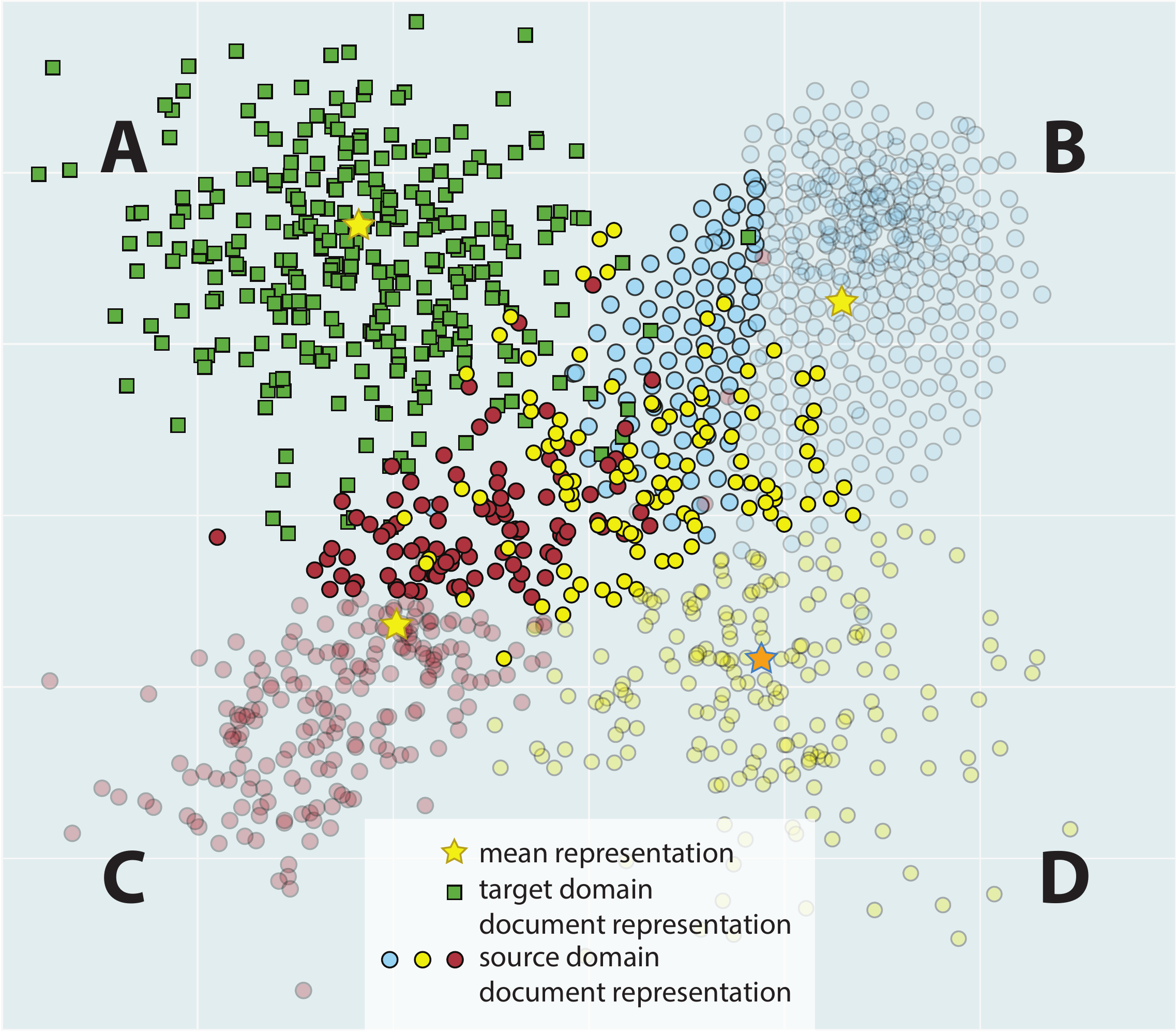}
    \caption{Instance-level selection} \label{fig:instance_level}
    \end{subfigure}
    \hspace*{0.1cm}
    \begin{subfigure}{.32\linewidth}
      \centering
         \includegraphics[width=0.9\linewidth]{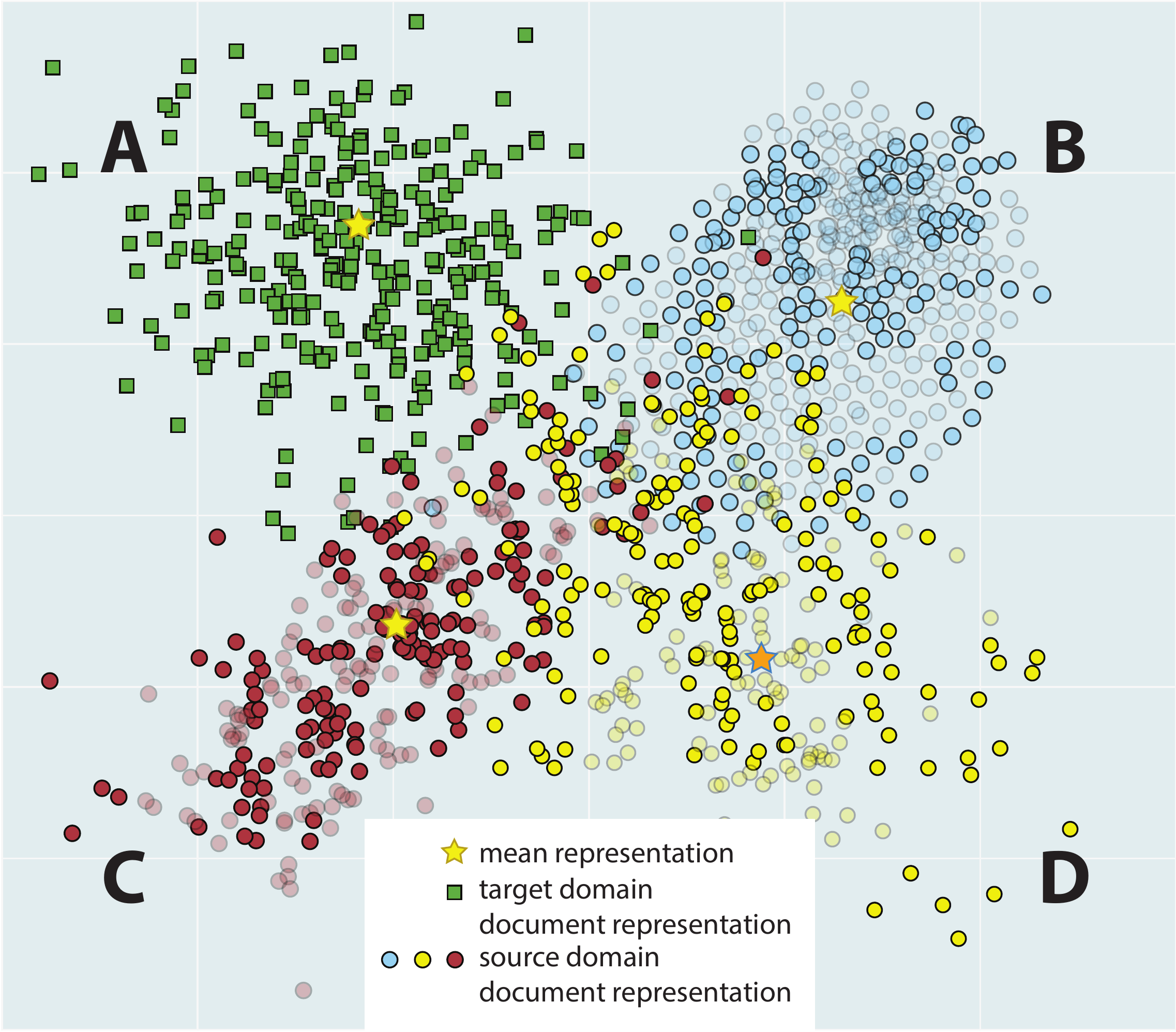}
    \caption{Subset-level selection} \label{fig:subset_level}
    \end{subfigure}
    \caption{Different data selection levels. Target domain examples are indicated by green squares and domain representations (cluster centroids) are indicated by yellow stars. The examples selected by each strategy are marked as solid. Subset-level selection leads to a more diverse training set.}
\label{fig:data_selection-strategies}
\end{figure*}

\textbf{Jensen-Shannon divergence.}  $\:$ Jensen-Shannon divergence is one of the most frequently used measures of domain similarity \cite{Remus2012} and has been shown to outperform other similarity metrics \cite{Plank2011}. Jensen-Shannon divergence is a smoothed, symmetric variant of KL divergence. The Jensen-Shannon divergence between two different probability distributions $P$ and $Q$ can be written as $D_{JS}(P||Q) = \frac{1}{2} [D_{KL}(P||M) + D_{KL}(Q||M)]$ where $M = \frac{1}{2} (P + Q)$, i.e. the average distribution of $P$ and $Q$, and $D_{KL}$ is the KL divergence:

\begin{equation}
D_{KL}(P||Q) = \sum_{i=1}^n p_i \frac{p_i}{q_i}.
\end{equation}

\textbf{Cosine similarity.} $\:$ Cosine similarity is traditionally used to measure the similarity between vectors, in particular word embeddings \cite{Mikolov2013d}. The cosine similarity between two vectors $a$ and $b$ is:

\begin{equation}
\text{cos}(a, b) = \dfrac{a \cdot b}{\|a\| \: \|b\|}.
\end{equation}

\textbf{Proxy $\mathcal{A}$ distance.} The $\:$ $\mathcal{A}$ distance \cite{Ben-David2007} aims to identify the subset $A$ in a family of subsets $\mathcal{A}$ on which the source domain distribution $P$ and the target domain distribution $Q$ differ the most and is defined as follows:

\begin{equation}
d_\mathcal{A} (P, Q) = 2 \: \sup\limits_{A \in \mathcal{A}} \:  | P(A) - Q(A) |.
\end{equation}

In practice, the Huber loss is used as a proxy for the $\mathcal{A}$ distance \cite{Blitzer2007} but is generally only used on a domain level to gain insights with regard to the adaptability of representations. We propose to use the proxy $\mathcal{A}$ distance as a data selection metric in the multi-domain setting: We first sample as many examples from the source domains as we have target domain examples. We then label all source domain examples with $0$ and all target domain examples with $1$ and train a logistic regression classifier on the balanced binary dataset. We then use the probability of belonging to the target domain as the similarity score for each source domain example.

\textbf{Other similarity metrics.} $\:$ Another similarity metric that has been successfully used in Machine Translation is a sentence's perplexity as determined by a language model trained on the target domain \cite{Duh2013}. However, in our experiments, perplexity proved unsuitable for data selection for sentiment analysis.

\subsection{Data selection level}

Finally, domain similarity can be measured on different levels as can be seen in Figure \ref{fig:data_selection-strategies}:  on the level of the entire domain (\ref{fig:domain_level}); on the level of a single training example (\ref{fig:instance_level}); or on a level that mediates between the two extremes and introduces diversity in the data selection (\ref{fig:subset_level}).   

\textbf{Domain level.} $\:$ Often, such as in the case of product reviews, domains are clearly delimited and documents in different domains are clearly distinct from one another. We can thus adopt these human-assigned labels and compute the similarity of each source domain with regard to the target domain. We then sample $n$ training examples only from the most similar source domain.

\textbf{Training instance level.} $\:$ In other scenarios, such as on the web, there is no clear distinction between different domains \cite{Ruder2016}. In this case, the similarity with regard to the target domain can be computed for each training instance \cite{Plank2011}. Instances are then sorted by similarity and the $n$ training examples with the highest similarity score are chosen for training.

\textbf{Instance subset level.} $\:$ While representations such as word embeddings have been shown to be effective at representing words or individual documents, term distributions are more apt to capture the statistics of a collection of examples, as the sparse term distribution of a single sentence or document might not contain sufficient evidence to compute an accurate notion of similarity on an instance level. To counter-act this, instead of calculating the similarity for each training instance, we propose to compute the similarities for random subsets of instances. In addition, considering a subset of examples in conjunction has the advantage of diversifying the training data and thus making the trained model more robust.

At each iteration, we sample $m$ subsets of size $s$ from all source domains $X$. We retain the subset with the highest similarity with regard to the target domain at each iteration and repeat the process until we have gathered $n$ training examples. The complete procedure is shown in Algorithm \ref{algo:subset_select}.
 
\begin{algorithm}
\caption{Instance subset select}\label{algo:subset_select}
\begin{algorithmic}[1]
\Procedure{SubsetSelect}{$s$, $n$, $m$, $X$}
\State $numiter \gets n / s$
\State $trainset \gets [\:]$
      \For{\texttt{i in range(numiter)}}
      
        \State $randsubsets \gets$ sample \emph{m} subsets of size $s$ from $X$
        \State $simscores \gets$ compute similarity score for $randsubsets$
        \State $sortedsimscores \gets$ sort $simscores$
        \State $bestsubset \gets$ get subset with highest similarity score
        \State $trainset$.append$(bestsubset)$
        \State $examples$.remove$(bestsubset)$
      \EndFor
\State \Return $trainset$
\EndProcedure
\end{algorithmic}
\end{algorithm}

\section{Experiments}

\subsection{Datasets and tasks}

As challenges for data selection methods vary depending on the nature of the domains to which they are applied, we evaluate our approaches on two large multi-domain sentiment analysis datasets, which seek to replicate these challenges. 

\textbf{Tweets+Reviews.} \: When data selection is applied to documents on the web, methods need to be able to handle domains that are not clearly defined and inconsistent. In the extreme, domains may be unknown or non-existing. In order to emulate this diversity, we choose the data of different editions of the SemEval Twitter Sentiment Analysis task as tweets within one domain are often heterogeneous \cite{Hovy2014}. Specifically, we select the training data of SemEval-2013 Task 9\footnote{\url{http://alt.qcri.org/semeval2014/task9/data/uploads/semeval2013_task2_train.zip}} (Twitter2013-Train) and the training (Twitter2016-Train) and test data of SemEval-2016 Task 4 Subtask A\footnote{\url{http://alt.qcri.org/semeval2016/task4/index.php?id=data-and-tools}}. We split the latter into its sub-domains, i.e. LiveJournal2014, SMS2013, Twitter2013, Twitter2014, Twitter2014Sarcasm, Twitter2015, Twitter2016. Furthermore, in order to gauge the methods' aptitude to handle diverse datasets, we include the laptop and restaurant reviews of SemEval-2016 Task 5\footnote{\url{http://alt.qcri.org/semeval2016/task5/index.php?id=data-and-tools}}.\footnote{We omit sentences with conflicting sentiments.} Statistics of the domains can be found in Table \ref{tab:tweets-stats}.

\begin{table}[]
\centering
\begin{tabular}{l c c}
\textbf{Dataset} & \textbf{\# of} & \textbf{\# of} \\
& \textbf{sentences} & \textbf{words} \\ \hline
LiveJournal2014 & 1142 & 15216 \\
SMS2013 & 2093 & 31774 \\
Twitter2013 & 3813 & 74649 \\
Twitter2014 & 1853 & 36305 \\
Twitter2014Sarcasm & 86 & 1219 \\
Twitter2015 & 2390 & 46270 \\
Twitter2016 & 20632 & 404827 \\
Twitter2016-Train & 5443 & 105984 \\
Twitter2013-Train & 7947 & 156507 \\
Laptop-reviews & 2426 & 31626 \\
Restaurant-reviews & 2110 & 25419
\end{tabular}
\caption{Number of sentences and number of words for different Twitter and review domains.}
\label{tab:tweets-stats}

\hspace{2cm}

\begin{tabular}{l c c c}
\textbf{Dataset} & \textbf{+} & \textbf{-} & \textbf{?} \\\hline
apparel & 1000 & 1000 & 7252 \\
automotive & 584 & 152 & 0 \\
baby & 1000 & 900 & 2356 \\
beauty & 1000 & 493 & 1391 \\
books & 1000 & 1000 & 973194 \\
camera \& photo & 1000 & 999 & 5409 \\
cell phones \& service & 639 & 384 & 0 \\
\begin{tabular}[c]{@{}l@{}}computer \&\\ video games\end{tabular} & 1000 & 458 & 1313 \\
dvd & 1000 & 1000 & 122438 \\
electronics & 1000 & 1000 & 21009 \\
gourmet food & 1000 & 208 & 367 \\
grocery & 1000 & 352 & 1280 \\
health \& personal care & 1000 & 1000 & 5225 \\
jewelry \& watches & 1000 & 292 & 689 \\
kitchen \& housewares & 1000 & 1000 & 17856 \\
magazines & 1000 & 970 & 2221 \\
music & 1000 & 1000 & 172180 \\
musical instruments & 284 & 48 & 0 \\
office products & 367 & 64 & 0 \\
outdoor living & 1000 & 327 & 272 \\
software & 1000 & 915 & 475 \\
sports \& outdoors & 1000 & 1000 & 3728 \\
tools \& hardware & 98 & 14 & 0 \\
toys \& games & 1000 & 1000 & 11147 \\
video & 1000 & 1000 & 34180
\end{tabular}
\caption{Number of positive (+), negative (-), and unlabeled documents (?) for the different review domains of the large Amazon reviews dataset.}
\label{tab:reviews-stats}
\end{table}

\textbf{Reviews.} \: We furthermore consider the diametrally opposite setting of clearly distinct domains. For this scenario, we use the large version of the reviews dataset of Blitzer et al. \shortcite{Blitzer2006}. The dataset consists of 25 different review domains with varying data sizes. To enforce realistic conditions where annotation is expensive we employ only the provided datasets with small amounts of labeled data. We show statistics in Table \ref{tab:reviews-stats}.

\textbf{Tasks.} \: We evaluate our methods on the ternary sentence-level and binary review-level sentiment analysis task on the Tweets+Reviews and Reviews dataset respectively.

\subsection{Training details}

We pre-process tweets by replacing urls, user names, and hashtags. We remove stopwords and use a vocabulary of the 10,000 most frequent words across all domains. In line with previous work, we use the raw bag-of-words unigram/bigram features pre-processed with tf-idf as input to a linear SVM classifier \cite{Blitzer2006}. We use GloVe vectors \cite{Pennington2014} pre-trained on 42B tokens of the Common Crawl corpus\footnote{\url{http://nlp.stanford.edu/projects/glove/}} for our word embeddings. For the auto-encoder representations, we use a denoising auto-encoder with one hidden layer of $1000$ dimensions and a masking noise with a masking probability $p = 0.8$, which we train for $50$ epochs with the Adam optimizer \cite{Kingma2015}. For subset-level selection, we set the subset size $s = 20$ and the number of subsets $m = 20000$ as determined via a grid search over a range of values on Tweets+Reviews validation data. 

We limit the number of training examples to $n=2000$ and $n=1600$ for Tweets+Reviews and Reviews respectively, the latter in accordance with the conventions of Bollegala et al. \shortcite{Bollegala2011}. In all cases, we evaluate on all data in the target domain.

\subsection{Comparison methods}

\textbf{Baselines.} \: We compare against two baselines: a) We randomly sample training examples from all source domains (\emph{rand}); b) we randomly sample the same number of stratified examples from all source domains (\emph{all}). Since the Reviews corpus contains distinct product review domains, we also include a human-labeled baseline (\emph{H}), where the model is trained on data randomly sampled from the domain that was determined most similar by 5 human annotators. We provided them with the names of the review categories and review samples and tasked them with ranking the three most similar domains for each domain, with the consensus being selected as the most similar domain.

\textbf{Our methods.} \: For each representation, we use -- due to space considerations -- the similarity metric that is most commonly used in conjunction with it, i.e. term distribution and Jensen-Shannon divergence, word embeddings and cosine similarity, and autoencoder representations and cosine similarity. For each combination, we select examples based on three levels: a) We randomly sample from the most similar domain ($\mathcal{D}$); b) we choose the most similar individual examples (\emph{ex}); and c) we choose the most similar subsets of examples (\emph{subset}). Due to space considerations, we apply proxy $\mathcal{A}$ distance for each representation only on the instance level, although it can be naturally combined with our subset-level selection strategy.

\subsection{Results}

For all datasets, we report accuracy scores and the average of 10 runs. We measure statistical significance using Student's T test.  We provide results in Tables \ref{tab:tweets-results} and \ref{tab:reviews-results}.

\begin{table*}
\centering
\resizebox{\textwidth}{!}{%
\begin{tabular}{l c c >{\columncolor[gray]{0.8}}c >{\columncolor[gray]{0.8}}c >{\columncolor[gray]{0.8}}c >{\columncolor[gray]{0.8}}c c c c c >{\columncolor[gray]{0.8}}c >{\columncolor[gray]{0.8}}c >{\columncolor[gray]{0.8}}c >{\columncolor[gray]{0.8}}c}
\textbf{Representation $\rightarrow$} & \multicolumn{2}{c}{} &  \multicolumn{4}{c}{{\cellcolor[gray]{.8}} \textbf{Term distribution}} & \multicolumn{4}{c}{\textbf{Word embeddings}} & \multicolumn{4}{c}{{\cellcolor[gray]{.8}}\textbf{AE representations}} \\\hline
\textbf{Metric $\rightarrow$} & \multicolumn{2}{c}{} & \multicolumn{3}{c}{{\cellcolor[gray]{.8}} \textbf{Jensen-Shannon}} & $\mathbf{d_\mathcal{A}}$ & \multicolumn{3}{c}{\textbf{Cosine similarity}} & $\mathbf{d_\mathcal{A}}$ & \multicolumn{3}{c}{{\cellcolor[gray]{.8}}\textbf{Cosine similarity}} & $\mathbf{d_\mathcal{A}}$ \\\hline
\textbf{Target domain $\downarrow$} & \emph{rand} & \emph{all} & $\mathcal{D}$ & \emph{ex} & \emph{subset} & \emph{ex} & $\mathcal{D}$ & \emph{ex} & \emph{subset} & \emph{ex} & $\mathcal{D}$ & \emph{ex} & \emph{subset} & \emph{ex} \\\hline
LiveJournal2014 & 50.9 & 50.6 & \bf 55.3$^{*\dagger}$ & 51.4 & 52.4$^{*\dagger}$ & 50.3 & 32.0 & 52.1$^{*\dagger}$ & 50.6 & 47.8 & 42.6 & 51.9 & 51.1 & 50.3 \\
SMS2013 & 53.6 & 47.7 & 57.0$^{*\dagger}$ & 53.9$^{\dagger}$ & 54.9$^\dagger$ & 53.0$^{\dagger}$ & \bf 60.8$^{*\dagger}$ & 56.0$^{*\dagger}$ & 51.2$^\dagger$ & 53.4$^*$ & 60.6$^{*\dagger}$ & 54.7$^{\dagger}$ & 51.7 & 52.7$^{\dagger}$ \\
Twitter2013 & 57.2 & 57.1 & 60.3$^{*\dagger}$ & 57.0 & 57.3 & 55.8 & 60.4$^{*\dagger}$ & 56.6 & 57.3 & 57.7 & \bf 61.0$^{*\dagger}$ & 56.6 & 56.8 & 58.3$^{*\dagger}$\\
Twitter2014 & 61.2 & 61.6 & \bf 63.2$^{*\dagger}$ & 60.8 & 61.8 & 61.5 & 62.1 & 62.1$^*$ & 61.8 & 62.5$^*$ & 62.6$^{*}$ & 59.6 & 61.8 & 62.3$^{*}$ \\
Twitter2014Sarcasm & 44.8 & 46.6 & 46.6 & 47.7$^{*}$ & 47.1 & 51.5$^{*\dagger}$ & 36.5 & 46.5 & 45.6 & 48.3$^*$ & 48.4$^{*}$ & 45.3 & 41.9 & \bf 53.5$^{*\dagger}$ \\
Twitter2015 & 56.7 & 56.2 & \bf 58.3$^{*\dagger}$ & 55.6 & 57.0 & 53.5 & 58.0$^{*\dagger}$ & 55.7 & 57.5 & 56.3 & 58.0$^{*\dagger}$ & 54.4 & 56.2 & 57.0 \\
Twitter2016 & 53.8 & 52.6 & \bf 56.3$^{*\dagger}$ & 51.3 & 53.7$^\dagger$ & 53.4$^{\dagger}$ & 54.7$^{*\dagger}$ & 47.1 & 54.1$^\dagger$ & 55.4$^{*\dagger}$ & 54.9$^{*\dagger}$ & 52.2 & 54.2$^{\dagger}$ & 55.2$^{*\dagger}$ \\
Twitter2016-Train & 50.6 & 50.0 & 49.5 & 51.7$^{*\dagger}$ & 50.6 & 45.4  & 51.6$^{*\dagger}$ & \bf 52.0$^{*\dagger}$ & 51.1$^\dagger$ & 47.9 & 50.0 & 46.9 & 48.1 & 51.1$^{\dagger}$ \\
Twitter2013-Train & 57.0 & 55.8 & 56.6$^{\dagger}$ & 55.6 & 56.8$^\dagger$ & 56.4 & 58.4$^{*\dagger}$ & 54.8 & 56.1 & 56.5 & 56.5 & 55.2 & 56.5 & \bf 57.5$^{\dagger}$ \\
Laptop-reviews & 48.1 & 54.1 & \bf 69.5$^{*\dagger}$ & 55.0$^*$ & 53.6$^*$ & 65.2 & 48.8 & 50.4$^*$ & 52.7$^*$ & 61.1$^{*\dagger}$ & 48.9 & 63.6$^{*\dagger}$ & 53.2$^{*}$ & 63.5$^{*\dagger}$\\
Restaurant-reviews & 56.3 & 60.1 & 71.8$^{*\dagger}$ & 61.4$^{*\dagger}$ & 60.9$^*$ & 69.6$^{*\dagger}$ & 52.5 & 63.4$^{*\dagger}$ & 61.8$^*$ & 67.1$^{*\dagger}$ & \bf 71.9$^{*\dagger}$ & 70.1$^{*\dagger}$ & 63.9$^{*\dagger}$ & 67.7$^{*\dagger}$\\\hline
Average & 53.6 & 53.8 & \bf 58.6$^{*\dagger}$ & 54.7$^{*\dagger}$ & 55.1$^{*\dagger}$ & 56.0$^{*\dagger}$ & 52.3 & 54.3 & 54.5 & 55.9$^{*\dagger}$ & 55.9$^{*\dagger}$ & 55.5$^{*\dagger}$ & 54.1 & 57.2$^{*\dagger}$
\end{tabular}
}
\caption{Comparison of different representations and metrics for multi-domain adaptation of tweets and review datasets for ternary sentence-level sentiment analysis. For each target domain, all other domains are available as source domains. Training examples are limited to $2000$. Baselines are random selection (\emph{rand}) and stratified selection balanced across all domains (\emph{all}). Training examples are a) samples from the most similar domain according to the chosen similarity metric ($\mathcal{D}$), b) the most similar individual examples (\emph{ex}), or c) the most similar subsets of examples (\emph{subset}). $d_\mathcal{A}$ is the proxy $\mathcal{A}$ distance. $^*$ and $^\dagger$ indicates significantly better ($p < 0.05$)  than \emph{rand} and \emph{all} baseline respectively.}
\label{tab:tweets-results} 

\hspace{2cm}

\resizebox{\textwidth}{!}{%
\begin{tabular}{l c c c >{\columncolor[gray]{0.8}}c >{\columncolor[gray]{0.8}}c >{\columncolor[gray]{0.8}}c >{\columncolor[gray]{0.8}}c c c c c >{\columncolor[gray]{0.8}}c >{\columncolor[gray]{0.8}}c >{\columncolor[gray]{0.8}}c >{\columncolor[gray]{0.8}}c}
\textbf{Representation $\rightarrow$} & \multicolumn{3}{c}{} &  \multicolumn{4}{c}{{\cellcolor[gray]{.8}} \textbf{Term distribution}} & \multicolumn{4}{c}{\textbf{Word embeddings}} & \multicolumn{4}{c}{{\cellcolor[gray]{.8}}\textbf{AE representations}} \\\hline
\textbf{Metric $\rightarrow$} & \multicolumn{3}{c}{} & \multicolumn{3}{c}{{\cellcolor[gray]{.8}} \textbf{Jensen-Shannon}} & $\mathbf{d_\mathcal{A}}$ & \multicolumn{3}{c}{\textbf{Cosine similarity}} & $\mathbf{d_\mathcal{A}}$ & \multicolumn{3}{c}{{\cellcolor[gray]{.8}}\textbf{Cosine similarity}} & $\mathbf{d_\mathcal{A}}$ \\\hline
\textbf{Target domain $\downarrow$} & \emph{rand} & \emph{all} & \emph{H} & $\mathcal{D}$ & \emph{ex} & \emph{subset} & \emph{ex} & $\mathcal{D}$ & \emph{ex} & \emph{subset} & \emph{ex} & $\mathcal{D}$ & \emph{ex} & \emph{subset} & \emph{ex} \\\hline
apparel & 81.4 & 79.6 & 78.4 & 83.8$^{*\dagger}$ & 78.3 & 81.5$^{\dagger}$ & 78.4 & 74.6 & 78.9 & 79.3  & 81.3$^\dagger$ & \bf 84.2$^{*\dagger}$ & 79.3 & 81.5$^\dagger$ & 80.8$^{\dagger}$\\
automotive & 80.1 & 83.2 & 79.4 & 79.0 & 75.7 & 81.2$^{*}$ & 83.0$^*$ & 83.8$^*$ & 82.1$^*$ & 83.0$^*$ & \bf 84.2$^{*\dagger}$ & 78.3 & 81.0 & 81.3$^*$ & 81.9$^{*}$ \\
baby & 80.0 & 78.7 & 81.2 & 80.9$^{*\dagger}$ & 78.6 & 81.0$^{*\dagger}$ & 80.2$^\dagger$ & \bf 82.2$^{*\dagger}$ & 80.9$^{\dagger}$ & 79.5 & 81.1$^{*\dagger}$ & \bf 82.2$^{*\dagger}$ & 81.3$^{*\dagger}$ & 81.5$^{*\dagger}$ & 81.5$^{*\dagger}$ \\
beauty & 81.3 & 80.9 & 84.1 & 83.8$^{*\dagger}$ & 81.3 & 82.6$^{*\dagger}$ & 80.7 & \bf 84.7$^{*\dagger}$ & 79.8 & 82.2$^\dagger$ & 83.1$^{*\dagger}$ & \bf 84.7$^{*\dagger}$ & 82.6$^{*\dagger}$ & 83.1$^{*\dagger}$ & 83.3$^{*\dagger}$ \\
books & 73.2. & 71.4 & 72.0 & \bf 79.6$^{*\dagger}$ & 73.8$^\dagger$ & 75.1$^{*\dagger}$ & 75.3$^{*\dagger}$ & 72.4$^{\dagger}$ & 73.9$^{\dagger}$ & 73.6$^\dagger$ & 75.2$^{*\dagger}$ & 77.2$^{*\dagger}$ & 76.7$^{*\dagger}$ & 75.2$^{*\dagger}$ & 74.9$^{*\dagger}$ \\
camera \& photo & 81.6 & 81.0 & 83.0 & 82.8$^{*\dagger}$ & 79.2 & 81.7 & 78.3 & \bf 83.3$^{*\dagger}$ & 82.1$^{\dagger}$ & 81.0 & 81.3 & \bf 83.3$^{*\dagger}$ & 80.9 & 81.2 & 80.2 \\
cell phones \& service & 81.6 & 81.9 & 81.9 & 81.4 & 76.3 & 81.1 & 78.4 & 72.8 & 81.5 & 81.5 & 80.3 & 81.6 & 79.9 & \bf 83.3$^{*\dagger}$ & 81.3 \\
computer \& video games & 80.1 & 80.6 & 72.7 &75.2 & 79.4 & 82.0$^{*\dagger}$ & 80.1 & 72.6 & 79.9 & 81.3 & 78.9 & 79.3 & 79.8 & \bf 82.1$^{*\dagger}$ & 80.4 \\
dvd & 76.7 & 75.6 & 80.4 & 80.3$^{*\dagger}$ & 77.2$^\dagger$ & 78.7$^{*\dagger}$ & 79.7$^{*\dagger}$ & \bf 81.1$^{*\dagger}$ & 76.8$^{\dagger}$ & 78.3$^{*\dagger}$ & 79.5$^{*\dagger}$ & \bf 81.1$^{*\dagger}$ & 80.6$^{*\dagger}$ & 79.2$^{*\dagger}$ & 79.0$^{*\dagger}$ \\
electronics & 79.4 & 78.2 & 68.3 & 82.0$^{*\dagger}$ & 75.5 & 80.5$^{*\dagger}$ & 79.8$^{\dagger}$ & \bf 82.4$^{*\dagger}$ & 79.6$^{\dagger}$ & 79.8$^\dagger$ & 81.5$^{*\dagger}$ & 79.1 & 80.9$^{*\dagger}$ & 80.3$^{*\dagger}$ & 81.3$^{*\dagger}$ \\
gourmet food & 82.3 & 84.4 & 86.8 & 86.8$^{*\dagger}$ & 76.3 & 83.4$^{*}$ & 86.8$^{*\dagger}$ &  86.8$^{*\dagger}$ & 83.2 &  84.3 & 85.7$^{*\dagger}$ & 86.8$^{*\dagger}$ & \bf 87.3$^{*\dagger}$ & 85.7$^{*\dagger}$ & 86.3$^{*\dagger}$ \\
grocery & 83.5 & 84.2 & 79.9 & 79.9 & 78.5 & 84.5$^*$ & 80.8 & 79.9 & 83.1 & 84.4 & \bf 86.2$^{*\dagger}$ & 79.9 & 85.7$^{*\dagger}$ & 83.8 & 85.1$^{*\dagger}$ \\
health \& personal care & 80.2 & 78.9 & 81.3 & 80.6$^{\dagger}$ & 76.7 & 79.5 & 76.7 & 81.2$^{*\dagger}$ & 80.6$^{\dagger}$ & 79.3 & 77.9 & \bf 81.9$^{*\dagger}$ & 80.7$^{\dagger}$ & 79.8 & 76.7 \\
jewelry \& watches & 84.7 & 86.2 & 82.9 & 82.9 & 80.7 & 86.2$^*$ & 87.5$^{*\dagger}$ & 83.7 & 83.5 & 86.9$^*$ & 86.3$^{*}$ & 84.4 & 86.1$^{*}$ & 87.5$^{*\dagger}$ & \bf 88.6$^{*\dagger}$ \\
kitchen \& housewares & 81.4 & 79.8 & 58.6 & 82.6$^{*\dagger}$ & 79.8 & 81.4$^\dagger$ & 72.6 &  74.3 & 82.0$^{\dagger}$ & 79.8 & 80.1 & \bf 83.2$^{*\dagger}$ & 81.6$^{\dagger}$ & 82.4$^{\dagger}$ & 78.9 \\
magazines & 75.9 & 75.1 & 75.8 & 74.8 & 76.5$^\dagger$ & \bf 78.9$^{*\dagger}$ & 73.6 & 75.0 & 75.7 & 74.3 & 75.8 & 75.8 & 77.6$^{*\dagger}$ & 76.4$^{\dagger}$ & 76.4$^{\dagger}$ \\
music & 73.4 & 71.0 & 50.1 & 78.1$^{*\dagger}$ & 73.1$^\dagger$ & 74.3$^{*\dagger}$ & 74.8$^{*\dagger}$ & \bf 78.9$^{*\dagger}$ & 72.0 & 73.7$^\dagger$ & 76.3$^{*\dagger}$ & \bf 78.9$^{*\dagger}$ & 71.7 & 74.5$^{*\dagger}$ & 74.3$^{\dagger}$ \\
musical instruments & 85.3 & 86.7 & 70.0 & 84.0 & 75.3 & 85.8 & 87.7$^{*}$ & 84.3 & 85.8 & 86.2 &  \bf 89.2$^{*\dagger}$ & 82.2 & 84.3 & 88.1$^{*\dagger}$ & 86.7$^{*}$ \\
office products & 82.0 & 83.4 & 85.2 & 82.7 & 72.4 & 81.9 & 85.2$^{*\dagger}$ & 85.4$^{*\dagger}$ & \bf 85.8$^{*\dagger}$ & 84.2$^{*}$ & \bf 85.8$^{*\dagger}$ & 77.5 & 79.8 & 83.1 & 83.8$^{*}$ \\
outdoor living & 81.9 & 84.0 & 82.0 & 80.2 & 74.4 & 81.8 & 82.7 & 81.6 & \bf 84.3$^{*}$ & 84.1$^*$ &  82.7 & 81.6 & 82.8 & 83.7$^{*}$ & 82.7  \\
software & 80.9 & 79.8 & 71.4 & 81.9$^{*\dagger}$ & 78.4 & 81.4$^\dagger$ & 82.0$^{*\dagger}$ &  \bf 83.2$^{*\dagger}$ & 81.2$^{\dagger}$ & 80.9 & 82.2$^{*\dagger}$ & \bf 83.2$^{*\dagger}$ & 79.3 & 82.4$^{*\dagger}$ & 82.6$^{*\dagger}$ \\
sports \& outdoors & 80.9 & 80.6 & 73.7 & 82.1$^{*\dagger}$ & 79.6 & 82.3$^{*\dagger}$ & 71.0 &  82.6$^{*\dagger}$ & 81.5 & 81.4 & 80.4 & \bf 83.8$^{*\dagger}$ & 80.7 & 82.2$^{*\dagger}$ & 79.2 \\
tools \& hardware & 80.5 & 85.2 & 82.3 & 77.5 & 80.4 & 85.0$^*$ & 85.7$^{*}$ & 91.1$^{*\dagger}$ & 78.6 & 83.0$^*$ & 85.7$^{*}$ & 76.8 & 85.4$^{*}$ & 85.7$^{*}$ & \bf 89.3$^{*\dagger}$ \\
toys \& games & 79.9 & 78.9 & 69.6 & 82.0$^{*\dagger}$ &78.3 & 79.8$^\dagger$ & 79.0 & \bf 82.8$^{*\dagger}$ & 78.9 & 79.1 & 80.9$^{*\dagger}$ & \bf 82.8$^{*\dagger}$ & 82.1$^{*\dagger}$ & 79.8 & 82.0$^{*\dagger}$ \\
video & 75.2 & 74.2 & 80.8 & 80.8$^{*\dagger}$ & 76.3$^{*\dagger}$ & 76.1$^{*\dagger}$ & 80.5$^{*\dagger}$ & \bf 81.3$^{*\dagger}$ & 71.4 &  77.4$^{*\dagger}$ & 79.1$^{*\dagger}$ & 81.3$^{*\dagger}$ & 79.8$^{*\dagger}$ & 78.3$^{*\dagger}$ & 79.8$^{*\dagger}$ \\\hline
Average & 80.2 & 80.1 & 76.1 & 81.0$^{*\dagger}$ & 77.3 & 81.1$^{*\dagger}$ & 80.2 & 80.9 & 80.1 & 80.7 & 81.6 & 81.2$^{*\dagger}$ & 81.1$^{*\dagger}$ & \bf 81.7$^{*\dagger}$ & 81.5$^{*\dagger}$
\end{tabular}
}
\caption{Comparison of different representations and metrics on multi-domain adaptation for document-level binary sentiment analysis. For each target domain, all other domains are available as source domains. Training examples are limited to $1600$. \emph{H} is data selection based on the most similar domain assigned by human annotators. For the rest of the legend, see Table \ref{tab:tweets-results}.}
\label{tab:reviews-results}
\end{table*}

\subsection{Discussion}

\textbf{Tweets+Reviews.} $\:$ Most methods outperform the random and balanced baselines on average. Term-distribution based domain-level selection achieves the best performance. Even though Twitter datasets are highly heterogeneous, training and test sets of the same edition of the competition have been collected in the same time frame and thus share similar topics and characteristics that are helpful for identifying the sentiment.

In contrast, the scores of the other selection strategies are similar across representations. Subset-level selection generally outperforms its instance-level counterpart. Using proxy $\mathcal{A}$ distance as a similarity metric generally improves upon using the default similarity metric for the given representation. Particularly for term distributions, it helps to mitigate the sparsity of the representation.

Even though domain-level selection using word embeddings outperforms the baselines on the majority of domains, they perform -- on average -- more poorly. This shows the fallacy of choosing the most similar domain for training: The domain might be similar with regard to the representations of the words that are used, but the way the words are used are different, as can be seen with the sarcasm, laptops, and restaurant domains, where semantically similar domains are chosen by word embeddings that are, however, less helpful for predicting sentiment. This also reveals that the mean of document representations might not be the best way to capture all aspects of domain similarity.

Autoencoder representations perform comparable to term distributions on the Tweets+Reviews dataset, while their domain-level selection method performs worse. We generally expected them to perform better than term distributions, but reason that this is mainly due to the scarcity of data in this multi-domain dataset.

\begin{figure}[!htb]
      \centering
         \includegraphics[width=\linewidth]{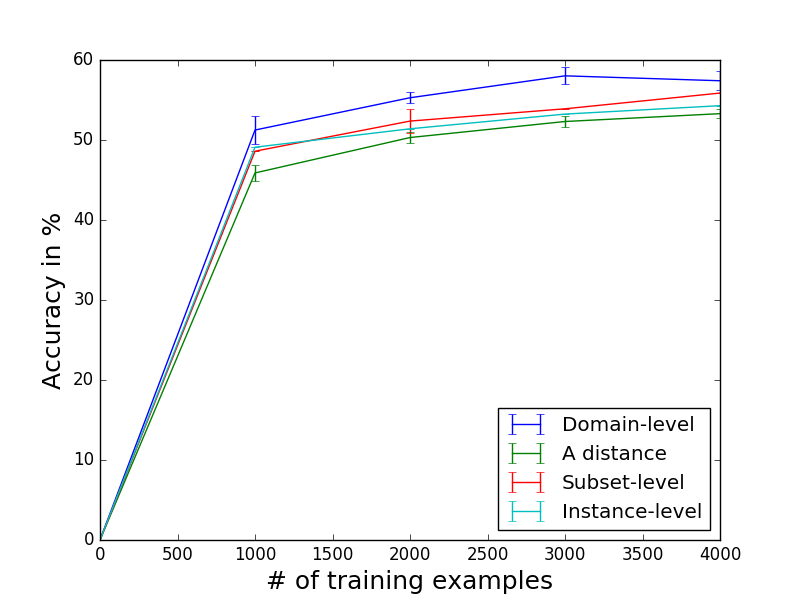}
    \caption{Comparison of average accuracy scores of different data selection methods on the LiveJournal2014 domain on the Tweets+Reviews with term distribution representation as representation and an increasing number of training examples.}
\label{fig:increasing_train_examples}
\end{figure}

As we have limited the number of training examples to $2000$, we also investigate the behaviour of the different data selection strategies when the number of training examples increases. We display this exemplarily for the LiveJournal2014 domain in Figure \ref{fig:increasing_train_examples}. We observe that -- while all selection strategies increase their performance -- subset-level selection improves slightly more considerably, but note that this trend might vary depending on the domain.

\textbf{Reviews.} $\:$ For the clearly distinct review domains, both the random and balanced baselines obtain comparatively stronger results. We attribute this to the fact that it is very difficult for a model to learn from a sentence that is not relevant to the target domain as in the Tweets+Reviews dataset, while an unhelpful review might still shed light on general sentiment words; in addition, the binary setting is easier than the ternary setting.

Domain-level selection based on term distributions still performs strongly, but is outperformed by subset-level selection on the same representations, which significantly improves upon instance-level selection. In comparison to the previous dataset, word embedding-based representation outperform term distributions, which we attribute to the fact that reviews are larger and contain more topical words with meaningful word representations in comparison to the noisy social media messages.

Even though product review categories have natural-language names, identifying the domain that is most similar and thus most likely to help the prediction at test time is no trivial task as can be seen with the abysmal performance of our human-labeled baseline (\emph{H}) and underlines the need for automatic data selection for domain adaptation. This is most evident with regard to the music domain, which is naturally conceptually similar to the musical instruments domain, but employs an entirely different repertoire of sentiment words. 

Another behavior that we observe is that while domain-level selection yields more often the best score for a domain compared to other selection strategies, its failure mode is equally polarized and often yields one of the worst scores for a target domain, if a source domain is matched whose examples are only peripherally relevant for the prediction.

Proxy $\mathcal{A}$ distance again consistently improves upon the default domain similarity metric demonstrating that domain similarity as judged by the confidence value of a liner classifier is a suitable metric for data selection. Finally, autoencoder representations outperform all other representations, while subset-level selection based on autoencoder representations performs best. The multiplicity of domains compensates for the lack of data, which renders autoencoder representations more meaningful and gives them an edge over term representations.

\subsection{Guidelines}

Finally, we propose guidelines on using data selection for sentiment analysis in the wild that might be helpful for NLP practitioners:

\begin{itemize}
	\item \textbf{DO} use term distribution-based domain-level selection with Jensen-Shannon divergence as a simple baseline. 
	\item \textbf{DO} use subset-level selection instead of instance-level selection, particularly with term distributions.
	\item \textbf{DO} use proxy $\mathcal{A}$ distance instead of the default similarity metric.
	\item \textbf{DO NOT} use pre-trained word embeddings for data selection on noisy data.
	\item \textbf{DO NOT} use autoencoder representations for data selection if your dataset is small.
	\item \textbf{DO} use autoencoder representations on large datasets.

\end{itemize}

\section{Conclusion}

In this paper, we have extensively studied -- for the first time as far as we are aware -- domain similarity metrics in the context of sentiment analysis. We have proposed several representations and metrics that have previously not been employed for data selection. We have introduced a novel data selection strategy that leverages subsets of examples and consistently outperforms instance-level selection. Finally, we have evaluated the proposed methods on two large-scale multi-domain adaptation settings on tweets and reviews and demonstrated that our proposed metrics and representations outperform strong random and balanced baselines, while our new selection strategy based on autoencoder representations yields the best score on the review corpus.

%\section*{Acknowledgments}
%
%Do not number the acknowledgment section.

\bibliography{data_selection}

\begin{thebibliography}{}
\expandafter\ifx\csname natexlab\endcsname\relax\def\natexlab#1{#1}\fi

\bibitem[{Ben-David et~al.(2007)Ben-David, Blitzer, Crammer, and
  Pereira}]{Ben-David2007}
Shai Ben-David, John Blitzer, Koby Crammer, and Fernando Pereira. 2007.
\newblock {Analysis of representations for domain adaptation}.
\newblock {\em Advances in Neural Information Processing Systems\/}
  19:137--144.

\bibitem[{Blitzer et~al.(2007)Blitzer, Dredze, and Pereira}]{Blitzer2007}
John Blitzer, Mark Dredze, and Fernando Pereira. 2007.
\newblock \href{https://doi.org/10.1109/IRPS.2011.5784441}{{Biographies,
  bollywood, boom-boxes and blenders: Domain adaptation for sentiment
  classification}}.
\newblock {\em Annual Meeting-Association for Computational Linguistics\/}
  45(1):440.
\newblock
  \href{https://doi.org/10.1109/IRPS.2011.5784441}{https://doi.org/10.1109/IRPS.2011.5784441}.

\bibitem[{Blitzer et~al.(2006)Blitzer, McDonald, and Pereira}]{Blitzer2006}
John Blitzer, Ryan McDonald, and Fernando Pereira. 2006.
\newblock \href{https://doi.org/10.3115/1610075.1610094}{{Domain Adaptation
  with Structural Correspondence Learning}}.
\newblock {\em EMNLP '06 Proceedings of the 2006 Conference on Empirical
  Methods in Natural Language Processing\/} (July):120--128.
\newblock
  \href{https://doi.org/10.3115/1610075.1610094}{https://doi.org/10.3115/1610075.1610094}.

\bibitem[{Bollegala et~al.(2011)Bollegala, Weir, and Carroll}]{Bollegala2011}
Danushka Bollegala, David Weir, and John Carroll. 2011.
\newblock {Using Multiple Sources to Construct a Sentiment Sensitive Thesaurus
  for Cross-Domain Sentiment Classification}.
\newblock In {\em Proceedings of the 49th Annual Meeting of the Association for
  Computational Linguistics: Human Language Technologies-Volume 1\/}. pages
  132--141.

\bibitem[{Chattopadhyay et~al.(2012)Chattopadhyay, Sun, Ye, Panchanathan, Fan,
  and Davidson}]{Chattopadhyay2012}
Rita Chattopadhyay, Qian Sun, Jieping Ye, Sethuraman Panchanathan, Wei Fan, and
  Ian Davidson. 2012.
\newblock {Multi-Source Domain Adaptation and Its Application to Early
  Detection of Fatigue}.
\newblock {\em ACM Transactions on Knowledge Discovery from Data (TKDD)\/}
  6(4).

\bibitem[{Chen et~al.(2012)Chen, Xu, Weinberger, and Sha}]{Chen2012}
Minmin Chen, Zhixiang Xu, Kilian~Q. Weinberger, and Fei Sha. 2012.
\newblock \href{https://doi.org/10.1007/s11222-007-9033-z}{{Marginalized
  Denoising Autoencoders for Domain Adaptation}}.
\newblock {\em Proceedings of the 29th International Conference on Machine
  Learning (ICML-12)\/} pages 767----774.
\newblock
  \href{https://doi.org/10.1007/s11222-007-9033-z}{https://doi.org/10.1007/s11222-007-9033-z}.

\bibitem[{{Daum{\'{e}} III}(2007)}]{DaumeIII2007a}
Hal {Daum{\'{e}} III}. 2007.
\newblock \href{https://doi.org/10.1.1.110.2062}{{Frustratingly Easy Domain
  Adaptation}}.
\newblock {\em Association for Computational Linguistic (ACL)s\/}
  (June):256--263.
\newblock
  \href{https://doi.org/10.1.1.110.2062}{https://doi.org/10.1.1.110.2062}.

\bibitem[{Duan et~al.(2009)Duan, Tsang, Xu, and Chua}]{Duan2009}
Lixin Duan, Ivor~W. Tsang, Dong Xu, and Tat-Seng Chua. 2009.
\newblock {Domain Adaptation from Multiple Sources via Auxiliary Classifiers}.
\newblock In {\em Proceedings of the 26th Annual International Conference on
  Machine Learning\/}.

\bibitem[{Duh et~al.(2013)Duh, Neubig, Sudoh, and Tsukada}]{Duh2013}
Kevin Duh, Graham Neubig, Katsuhito Sudoh, and Hajime Tsukada. 2013.
\newblock {Adaptation Data Selection using Neural Language Models : Experiments
  in Machine Translation}.
\newblock {\em Acl-2013\/} (1):678--683.

\bibitem[{Glorot et~al.(2011)Glorot, Bordes, and Bengio}]{Glorot2011a}
Xavier Glorot, Antoine Bordes, and Yoshua Bengio. 2011.
\newblock \href{http://www.icml-2011.org/papers/342{\_}icmlpaper.pdf}{{Domain
  Adaptation for Large-Scale Sentiment Classification: A Deep Learning
  Approach}}.
\newblock {\em Proceedings of the 28th International Conference on Machine
  Learning\/} (1):513--520.
\newblock
  \href{http://www.icml-2011.org/papers/342{\_}icmlpaper.pdf}{http://www.icml-2011.org/papers/342{\_}icmlpaper.pdf}.

\bibitem[{Hovy et~al.(2014)Hovy, Plank, and S{\o}gaard}]{Hovy2014}
Dirk Hovy, Barbara Plank, and Anders S{\o}gaard. 2014.
\newblock
  \href{http://aclanthology.info/papers/when-pos-data-sets-don-t-add-up-combatting-sample-bias}{{When
  POS data sets don't add up: Combatting sample bias}}.
\newblock {\em Proceedings of the Ninth International Conference on Language
  Resources and Evaluation (LREC-2014)\/} pages 4472--4475.
\newblock
  \href{http://aclanthology.info/papers/when-pos-data-sets-don-t-add-up-combatting-sample-bias}{http://aclanthology.info/papers/when-pos-data-sets-don-t-add-up-combatting-sample-bias}.

\bibitem[{Jiang and Zhai(2007)}]{Jiang2007}
Jing Jiang and ChengXiang Zhai. 2007.
\newblock \href{https://doi.org/10.1145/1273496.1273558}{{Instance Weighting
  for Domain Adaptation in NLP}}.
\newblock {\em Proceedings of the 45th Annual Meeting of the Association of
  Computational Linguistics\/} (October):264--271.
\newblock
  \href{https://doi.org/10.1145/1273496.1273558}{https://doi.org/10.1145/1273496.1273558}.

\bibitem[{Kingma and Ba(2015)}]{Kingma2015}
Diederik~P. Kingma and Jimmy~Lei Ba. 2015.
\newblock {Adam: a Method for Stochastic Optimization}.
\newblock {\em International Conference on Learning Representations\/} pages
  1--13.

\bibitem[{Lu et~al.(2011)Lu, Ott, Cardie, and Tsou}]{Lu2011}
Bin Lu, Myle Ott, Claire Cardie, and Benjamin Tsou. 2011.
\newblock {Multi-aspect Sentiment Analysis with Topic Models}.
\newblock In {\em Data Mining Workshops (ICDMW), 2011 IEEE 11th International
  Conference\/}. IEEE, pages 81--88.

\bibitem[{Mansour(2009)}]{Mansour2009a}
Yishay Mansour. 2009.
\newblock {Domain Adaptation with Multiple Sources}.
\newblock {\em NIPS\/} .

\bibitem[{Mikolov et~al.(2013)Mikolov, Chen, Corrado, and Dean}]{Mikolov2013d}
Tomas Mikolov, Kai Chen, Greg Corrado, and Jeffrey Dean. 2013.
\newblock {Distributed Representations of Words and Phrases and their
  Compositionality}.
\newblock {\em NIPS\/} pages 1--9.

\bibitem[{Pan et~al.(2010)Pan, Ni, Sun, Yang, and Chen}]{Pan2010a}
Sinno~Jialin Pan, Xiaochuan Ni, Jian-tao Sun, Qiang Yang, and Zheng Chen. 2010.
\newblock {Cross-Domain Sentiment Classification via Spectral Feature
  Alignment}.
\newblock In {\em Proceedings of the 19th International Conference on World
  Wide Web\/}. pages 751--760.

\bibitem[{Pennington et~al.(2014)Pennington, Socher, and
  Manning}]{Pennington2014}
Jeffrey Pennington, Richard Socher, and Christopher~D. Manning. 2014.
\newblock \href{https://doi.org/10.3115/v1/D14-1162}{{Glove: Global Vectors for
  Word Representation}}.
\newblock {\em Proceedings of the 2014 Conference on Empirical Methods in
  Natural Language Processing\/} pages 1532--1543.
\newblock
  \href{https://doi.org/10.3115/v1/D14-1162}{https://doi.org/10.3115/v1/D14-1162}.

\bibitem[{Plank(2016)}]{Plank2016}
Barbara Plank. 2016.
\newblock \href{https://arxiv.org/pdf/1608.07836.pdf}{{What to do about
  non-standard (or non-canonical) language in NLP}}.
\newblock {\em KONVENS 2016\/}
  \href{https://arxiv.org/pdf/1608.07836.pdf}{https://arxiv.org/pdf/1608.07836.pdf}.

\bibitem[{Plank and van Noord(2011)}]{Plank2011}
Barbara Plank and Gertjan van Noord. 2011.
\newblock {Effective Measures of Domain Similarity for Parsing}.
\newblock {\em Proceedings of the 49th Annual Meeting of the Association for
  Computational Linguistics: Human Language Technologies\/} 1:1566--1576.

\bibitem[{Remus(2012)}]{Remus2012}
Robert Remus. 2012.
\newblock {Domain adaptation using Domain Similarity- and Domain
  Complexity-based Instance Selection for Cross-Domain Sentiment Analysis}.
\newblock In {\em IEEE ICDM SENTIRE-2012\/}.

\bibitem[{Ruder et~al.(2016)Ruder, Ghaffari, and Breslin}]{Ruder2016}
Sebastian Ruder, Parsa Ghaffari, and John~G Breslin. 2016.
\newblock {Towards a continuous modeling of natural language domains}.
\newblock In {\em Uphill Battles in Language Processing Workshop, EMNLP
  2016\/}.

\bibitem[{Ruder et~al.(2017)Ruder, Ghaffari, and Breslin}]{Ruder2017a}
Sebastian Ruder, Parsa Ghaffari, and John~G. Breslin. 2017.
\newblock {Knowledge Adaptation : Teaching to Adapt}.
\newblock In {\em arXiv preprint arXiv:1702.02052\/}.

\bibitem[{{Van Asch} and Daelemans(2010)}]{VanAsch2010}
Vincent {Van Asch} and Walter Daelemans. 2010.
\newblock \href{http://eprints.pascal-network.org/archive/00007014/}{{Using
  Domain Similarity for Performance Estimation}}.
\newblock {\em Computational Linguistics\/} (July):31--36.
\newblock
  \href{http://eprints.pascal-network.org/archive/00007014/}{http://eprints.pascal-network.org/archive/00007014/}.

\bibitem[{Wieting et~al.(2016)Wieting, Bansal, Gimpel, and
  Livescu}]{Wieting2016}
John Wieting, Mohit Bansal, Kevin Gimpel, and Karen Livescu. 2016.
\newblock \href{http://arxiv.org/abs/1511.08198}{{Towards Universal
  Paraphrastic Sentence Embeddings}}.
\newblock In {\em ICLR\/}.
\newblock
  \href{http://arxiv.org/abs/1511.08198}{http://arxiv.org/abs/1511.08198}.

\bibitem[{Wu and Huang(2016)}]{Wu2016a}
Fangzhao Wu and Yongfeng Huang. 2016.
\newblock {Sentiment Domain Adaptation with Multiple Sources}.
\newblock {\em Proceedings of the 54th Annual Meeting of the Association for
  Computational Linguistics (ACL 2016)\/} pages 301--310.

\bibitem[{Yang and Eisenstein(2015)}]{Yang2015d}
Yi~Yang and Jacob Eisenstein. 2015.
\newblock {Unsupervised Multi-Domain Adaptation with Feature Embeddings}.
\newblock {\em Naacl-2015\/} 1:503--513.

\bibitem[{Zhou et~al.(2016)Zhou, Xie, Huang, and He}]{Zhou2016}
Guangyou Zhou, Zhiwen Xie, Jimmy~Xiangji Huang, and Tingting He. 2016.
\newblock
  \href{https://www.aclweb.org/anthology/P/P16/P16-1031.pdf}{{Bi-Transferring
  Deep Neural Networks for Domain Adaptation}}.
\newblock {\em ACL\/} pages 322--332.
\newblock
  \href{https://www.aclweb.org/anthology/P/P16/P16-1031.pdf}{https://www.aclweb.org/anthology/P/P16/P16-1031.pdf}.

\bibitem[{Zhuang et~al.(2015)Zhuang, Cheng, Luo, Pan, and He}]{Zhuang2015}
Fuzhen Zhuang, Xiaohu Cheng, Ping Luo, Sinno~Jialin Pan, and Qing He. 2015.
\newblock {Supervised Representation Learning: Transfer Learning with Deep
  Autoencoders}.
\newblock {\em IJCAI International Joint Conference on Artificial
  Intelligence\/} pages 4119--4125.

\end{thebibliography}
\bibliographystyle{acl_natbib}

\end{document}